\pdfoutput=1

\documentclass[11pt]{article}

\usepackage{acl}

\usepackage{times}
\usepackage{latexsym}

\usepackage[T1]{fontenc}

\usepackage[utf8]{inputenc}

\usepackage{microtype}

\usepackage{inconsolata}

\usepackage{hyperref}       
\usepackage{url}            
\usepackage{booktabs}       
\usepackage{amsfonts}       
\usepackage{nicefrac}       
\usepackage{xcolor}         
\usepackage{multirow}
\usepackage{graphicx}
\usepackage{tcolorbox}
\usepackage{soul}

\newtheorem{observation}{Observation}%

\usepackage{xcolor}

\newcommand{\prompt}[1]{%
  \begin{tcolorbox}[%
    colback=gray!10,    
    colframe=gray,      
    fontupper=\small,     
    title= Prompt with placeholder     
  ]
    #1
  \end{tcolorbox}%
}

\newcommand{\qa}[1]{%
  \begin{tcolorbox}[%
    colback=blue!10,    
    colframe=blue,      
    fontupper=\small,     
    title=LLM generated question-answer pair  
  ]
    #1
  \end{tcolorbox}%
}

\newcommand{\eval}[1]{%
  \begin{tcolorbox}[%
    colback=orange!10,    
    colframe=orange,      
    fontupper=\small,     
    title=Evaluation comparison of LLM generated answer  
  ]
    #1
  \end{tcolorbox}%
}

\newcommand{\ragas}{$\mathsf{RAGAS}$~}

\usepackage{glossaries}
\newacronym{llm}{LLM}{Large Language Model}
\newacronym{rag}{RAG}{Retrieval Augmented Generation}
\newacronym{nlp}{NLP}{Natural Language Processing}
\newacronym{eis}{EIS}{Environmental Impact Statements}

\title{WeQA: A Benchmark for Retrieval Augmented Generation in Wind Energy Domain}
\author{%
  Rounak Meyur,
  Hung Phan,
  Sridevi Wagle,
  Jan Strube,
  Mahantesh Halappanavar,\\
  \textbf{Sameera Horawalavithana,
  Anurag Acharya,
  Sai Munikoti}\\
  Pacific Northwest National Laboratory\\
  Richland, WA 99354 \\
  \texttt{\{rounak.meyur, hung.phan, sridevi.wagle, jan.strube, mahantesh.halappanavar,}\\ \texttt{yasanka.horawalavithana, anurag.acharya, sai.munikoti\}@pnnl.gov}
}

\begin{document}

\maketitle

\begin{abstract}
Wind energy project assessments present significant challenges for decision-makers, who must navigate and synthesize hundreds of pages of environmental and scientific documentation.
These documents often span different regions and project scales, covering multiple domains of expertise.
This process traditionally demands immense time and specialized knowledge from decision-makers.
The advent of \gls*{llm}s and \gls*{rag} approaches offer a transformative solution, enabling rapid, accurate cross-document information retrieval and synthesis.
As the landscape of \gls*{nlp} and text generation continues to evolve, benchmarking becomes essential to evaluate and compare the performance of different \gls*{rag}-based \gls*{llm}s. 
In this paper, we present a comprehensive framework to generate a domain relevant \gls*{rag} benchmark. Our framework is based on automatic question-answer generation with Human (domain experts)-AI (\gls*{llm}) teaming. 
As a case study, we demonstrate the framework by introducing WeQA, a first-of-its-kind benchmark on the wind energy domain which comprises of multiple scientific documents/reports related to environmental aspects of wind energy projects. 
Our framework systematically evaluates RAG performance using diverse metrics and multiple question types with varying complexity level, providing a foundation for rigorous assessment of \gls*{rag}-based systems in complex scientific domains and enabling researchers to identify areas for improvement in domain-specific applications.
\end{abstract}

\section{Introduction}
In recent years, the advancements in \gls*{llm} have revolutionized various natural language processing tasks, including text and response generation.
However, text generation using \gls*{llm} often encounters challenges such as generating irrelevant or incoherent outputs, perpetuating biases ingrained in the training data, and struggling to maintain context and factual accuracy~\cite{wu2024surveyllmgeneratedtextdetection}. These issues pose significant obstacles to achieving human-level performance in automated text generation systems. \gls*{rag} effectively mitigates these common challenges by incorporating retrieved information to enhance coherence and factual accuracy, thus minimizing the generation of fictitious or irrelevant content~\cite{gao2024,lewis2021}. Furthermore, concurrent works suggest RAG is the most sought approach for adapting models towards accelerating repetitive and data intensive tasks in niche scientific domain such as nuclear, renewable energy, environmental policy, etc. \cite{munikoti2024atlantic, munikoti2024evaluating, phan2024rag}. 
While \gls*{rag}-based systems have demonstrated promising capabilities in streamlining document analysis tasks across various professional domains, their integration into critical decision-making processes like permitting wind energy projects remains constrained due to legitimate concerns about trust and reliability.

In this work, we create benchmarks to assess \gls*{rag}-based \gls*{llm} performance in the domain of permitting wind energy projects.
\gls*{eis} represent the cornerstone documentation within this permitting landscape, serving as comprehensive analyses that evaluate the potential environmental consequences of proposed wind energy developments. 
These documents play a pivotal role in promoting informed decision-making by ensuring transparency and incorporating diverse stakeholder perspectives into the approval process~\cite{eis}.
By providing detailed evaluations of environmental effects, alternatives analysis, and mitigation measures, \gls*{eis} documentation facilitates the responsible development of wind energy infrastructure while building public trust at the same time.

As \gls*{rag}-based \gls*{llm}s gain traction for domain-specific applications such as wind energy permitting, their effectiveness must be rigorously assessed through robust benchmarks to ensure its practical utility and reliability~\cite{chen2023}. 
Establishing high-quality benchmarks is essential to evaluate their abilities to perform regulatory-focused reasoning, accurately interpret complex \gls*{eis} documents, and support logical deductions grounded in the documents.
Such benchmarks facilitate systematic assessment of how well \gls*{rag}-based \gls*{llm}s can handle the nuanced requirements of the domain~\cite{xiong2024}. 
A robust evaluation framework allows researchers and practitioners to investigate the impact of retrieval strategies, model architectures, and training data, on the performance of RAG, while building confidence in automated tools for critical environmental decision making~\cite{ray2023}.

In benchmarking \gls*{rag} for wind energy project permitting applications, it is crucial to evaluate its performance across a diverse set of questions that reflect the complexity and variability of real-world permitting scenarios~\cite{lyu2024}. 
A set of well curated and diverse questions enable a comprehensive assessment of \gls*{rag}'s ability to interpret \gls*{eis} documents, analyze environmental impacts, evaluate regulatory compliance, and generate coherent responses to permitting-related queries that practitioners encounter during wind energy project review processes.
To generate such questions, automated methods leveraging \gls*{nlp} techniques can be employed, including rule-based approaches that capture language patterns from relevant documents, template filling methods that incorporate wind energy terminologies, and neural network-based models that can efficiently create diverse question sets by leveraging the semantic relationships inherent in \gls*{eis} and other documents related to wind energy projects. 

Human-curated questions offer a level of linguistic richness and contextual relevance that may be challenging to achieve solely through automated generation methods, particularly in specialized domains such as wind energy project permitting~\cite{zhang2024}. By leveraging human expertise and domain knowledge, curated question sets can encompass a broader spectrum of linguistic variations, domain-specific considerations, and nuanced semantics~\cite{ribeiro-etal-2020-beyond}, providing a more comprehensive evaluation of RAG's performance across diverse scenarios and applications~\cite{thakur2beir}. Combining automated generation with human curation for benchmarking \gls*{rag} offers a synergistic approach to ensure both efficiency and quality in question sets. This hybrid approach leverages the strengths of both automated and human-driven processes, that provide efficient and robust evaluation metrics for RAG's performance.

In this work, we present a hybrid workflow to benchmark \gls*{rag}s, which combines rapid question generation through automated methods, augmented with properly designed human prompts to generate diverse set of questions.
Our proposed benchmarking framework is used to generate questions from \gls*{eis} and other research documents related to environmental impact of wind energy projects. 
The extensive question-answer dataset serve as a tool to evaluate the performance of \gls*{rag}-based \gls*{llm}s, which are designed to answer queries related to these extensive and comprehensive documents. 
Given the vast amount of information contained in these documents, manually reviewing them is impractical, making \gls*{rag}-based \gls*{llm}s essential for generating accurate responses to specific queries. 
Our benchmarking framework assesses the effectiveness of these models in accurately retrieving and responding to queries, ensuring that they can reliably process and provide relevant information from the documents.

\noindent\textbf{Contributions}~~The paper introduces a novel benchmark dataset for question-answering (QA) task in a specific domain and also proposes a generic framework to evaluate the \gls*{rag}-based \gls*{llm} responses to different entries in the benchmark. 
This framework is designed to be adaptable across various domains, with a specific focus on documents related to wind energy project permitting in this study. 
The contributions of this research are as follows: 

\paragraph{Novel domain-specific benchmark.}~We present WeQA,\footnote{This benchmark will be made publicly available.} the first comprehensive benchmark QA dataset specifically designed for the wind energy domain, addressing the gap in specialized evaluation datasets for wind energy project permitting.

\paragraph{Domain-agnostic framework.}~Our proposed benchmark creation and \gls*{llm} evaluation framework is domain-agnostic and can be tailored for any desired niche domain, enabling researchers to adapt the methodology for various specialized fields beyond wind energy.

\paragraph{Hybrid question generation.}~We introduce a hybrid method that automatically generates diverse question types with varying complexity levels, producing both objective and subjective responses across different document sections to comprehensively evaluate \gls*{llm} performance.

\paragraph{Scalable evaluation methodology.}~We utilize established scoring frameworks like RAGAS~\cite{es2023ragas} and incorporate multiple \gls*{llm}s as judges, ensuring scalability, reproducibility, and comprehensive performance assessment of \gls*{rag}-based systems.



\section{Related Works}
\label{sec:related}

There have been a lot of work in the field of benchmarking, particularly for question answering (QA) task. These can be broadly divided into general QA and domain-specific QA.

\noindent\textbf{General QA benchmarks.}~
These benchmarks have established foundational evaluation frameworks for reading comprehension and knowledge retrieval tasks.
Notable general QA benchmarks include reading comprehension datasets such as the Stanford Question Answering Dataset (SQuAD)~\cite{rajpurkar2016squad} and MCTest~\cite{richardson2013mctest}, reasoning-focused benchmarks like the AI2 Reasoning Challenge (ARC)~\cite{clark2018think}, and comprehensive evaluation suites such as GLUE~\cite{wang2018glue} and Big Bench~\cite{srivastava2022beyond}. 
Additional benchmarks targeting open-domain knowledge include CommonsenseQA \cite{talmor2018commonsenseqa}, TriviaQA~\cite{joshi2017triviaqa}, Search QA~\cite{dunn2017searchqa}, and NewsQA~\cite{trischler2016newsqa}.

\noindent\textbf{Domain-specific QA benchmarks.}~
Recognizing the limitations of general benchmarks for specialized applications, researchers have developed domain-specific evaluation frameworks that capture the unique linguistic patterns, technical terminology, and reasoning requirements of particular fields. 
While scientific benchmarks such as MMLU~\cite{hendrycks2020measuring}, SciBench~\cite{wang2023scibench}, SciQ~\cite{welbl2017crowdsourcing}, SciRepEval~\cite{singh2022scirepeval}, SciQA~\cite{auer2023sciqa}, and QASA~\cite{lee2023qasa} are used for multi-disciplinary scientific QA evaluations, field-specific benchmarks include TheoremQA~\cite{chen2023theoremqa} for mathematics, emrQA~\cite{pampari2018emrqa} for medicine, BioRead~\cite{pappas2018bioread} and BioMRC~\cite{pappas2020biomrc} for biology, LawBench~\cite{Chen2023LawBenchBLE} for legal, and NuclearQA~\cite{acharya2023nuclearqa} for nuclear domains.

For environmental assessment specifically, benchmarks such as EnviroExam~\cite{huang2024enviroexam} for environmental science QA and NEPAQuAD~\cite{phan2024rag} for Environmental Impact Statement (EIS) documents have emerged.
However, to our knowledge, no benchmarks exist specifically for wind energy project permitting, making the proposed WeQA benchmark the first comprehensive benchmarking effort in this critical domain.
\begin{figure*}
    \centering
    \includegraphics[width=0.95\textwidth]{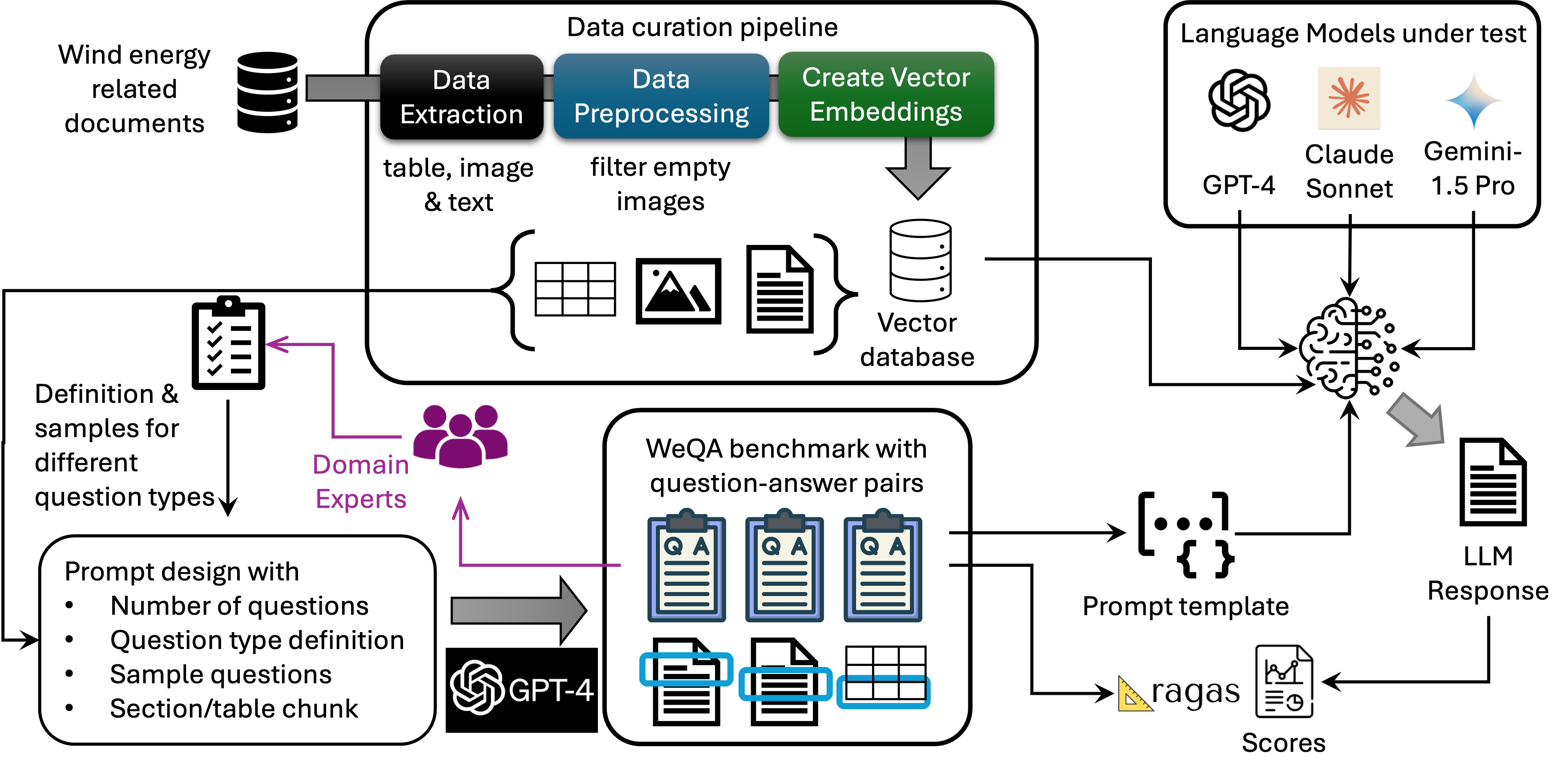}
    \caption{An overview of the proposed \gls*{rag} benchmarking framework. Multiple versions of hybrid questions are generated from specific text chunks of source documents with human-in-the-loop to review them. These questions are used as prompts for the \gls*{llm} or \gls*{rag} model under test.}
    \label{fig:framework}
\end{figure*}

\section{Dataset Creation}
In this paper, we focus on wind energy-related documents to enable the \gls*{rag}-based \gls*{llm}s to answer questions specific to this field. 
We gather PDF documents, including research articles and environmental impact studies published by the Department of Energy (DOE) under the National Environmental Policy Act (NEPA). Accessing information from this vast database is not straightforward, necessitating the need for a trained \gls*{llm} to accurately retrieve and answer questions from the provided context. The challenge is to ensure that the model's responses are based on the actual documents and do not hallucinate information. By using \gls*{rag}-based \gls*{llm}s, we aim to enhance the reliability and accuracy of responses related to wind energy, leveraging the rich information within our extensive document collection. This approach ensures that the information provided is both relevant and grounded in the sourced material.

We constructed a data extraction and curation pipeline to extract text, image, and table information from wind energy-related documents as depicted in the `data curation pipeline' in Figure~\ref{fig:framework}. Utilizing large language model (LLM) based methods such as the \emph{Unstructured.io} tool~\cite{unstructured}, we efficiently extracted information and converted it into JSON elements. 
To ensure data quality, we implemented a filtering step to remove images without meaningful content, such as decorative elements or blank spaces.
These filtered JSON elements were then organized into a schema, creating a page-wise assortment of text, table, and image elements. This structured format ensures that the extracted data is easily accessible and can be accurately referenced during model training and evaluation.



\section{Methodology}
While past works have generally preferred to use crowdsourcing as a way to craft datasets and benchmarks \cite{sap2019atomic, acharya2021towards}, we choose to automated methods for benchmark question generation. Automatically generating benchmarking questions using GPT-4 allows for efficient and scalable evaluation of other \gls*{llm}s and \gls*{rag}. However, this approach can introduce errors, leading to poor quality of questions being generated. This makes it essential to incorporate a human-in-the-loop for reviewing and refining the questions and responses. This paper proposes hybrid approaches, where automated methods are combined with human curation to ensure the accuracy and reliability of the benchmarking process. By leveraging both machine and human expertise, we can achieve more robust and comprehensive benchmarking framework.

Figure~\ref{fig:framework} provides an overview of the proposed \gls*{llm} benchmarking framework.
The core of the benchmarking framework is the question generation aspect, where automatic generation of questions forms the foundation. 
We combine this with human curation to select high-quality questions, ensuring relevance and clarity. 
Corresponding answers to these questions are then validated by humans, establishing a reliable ground truth. 
This curated set of questions and validated answers is used to evaluate the responses of other \gls*{llm}s and \gls*{rag} models. 


\noindent\textbf{Different question types.}~We generate multiple types of questions, including closed, open, comparison, evaluation, recall, process, and rhetorical questions. 
This diversity ensures a comprehensive benchmarking process, as each question type assesses different aspects of the models' capabilities. 
By incorporating a wide variety of questions, we can more effectively evaluate and compare the performance of \gls*{llm}s and \gls*{rag} models across various dimensions. 
This approach provides a holistic view of their strengths and weaknesses.

Each of these question type evaluates different capabilities of the \gls*{llm} under test.
\emph{Open questions} require models to generate detailed, free-form responses, testing their ability to construct coherent and informative answers.
\emph{Comparison questions} ask models to compare and contrast different concepts or entities, assessing their analytical and comparative reasoning skills.
\emph{Evaluation questions} require models to make judgments or provide assessments, gauging their ability to evaluate information critically.
\emph{Recall questions} focus on the model's ability to retrieve and reproduce specific information from memory, testing their factual accuracy.
\emph{Process questions} ask models to explain processes or sequences of actions, evaluating their understanding of procedures and logical progression.
\emph{Rhetorical questions} are used to test the models' grasp of nuances in language and their ability to recognize and appropriately respond to questions that may not require direct answers.
\begin{figure*}
    \centering
    \includegraphics[width=0.99\textwidth]{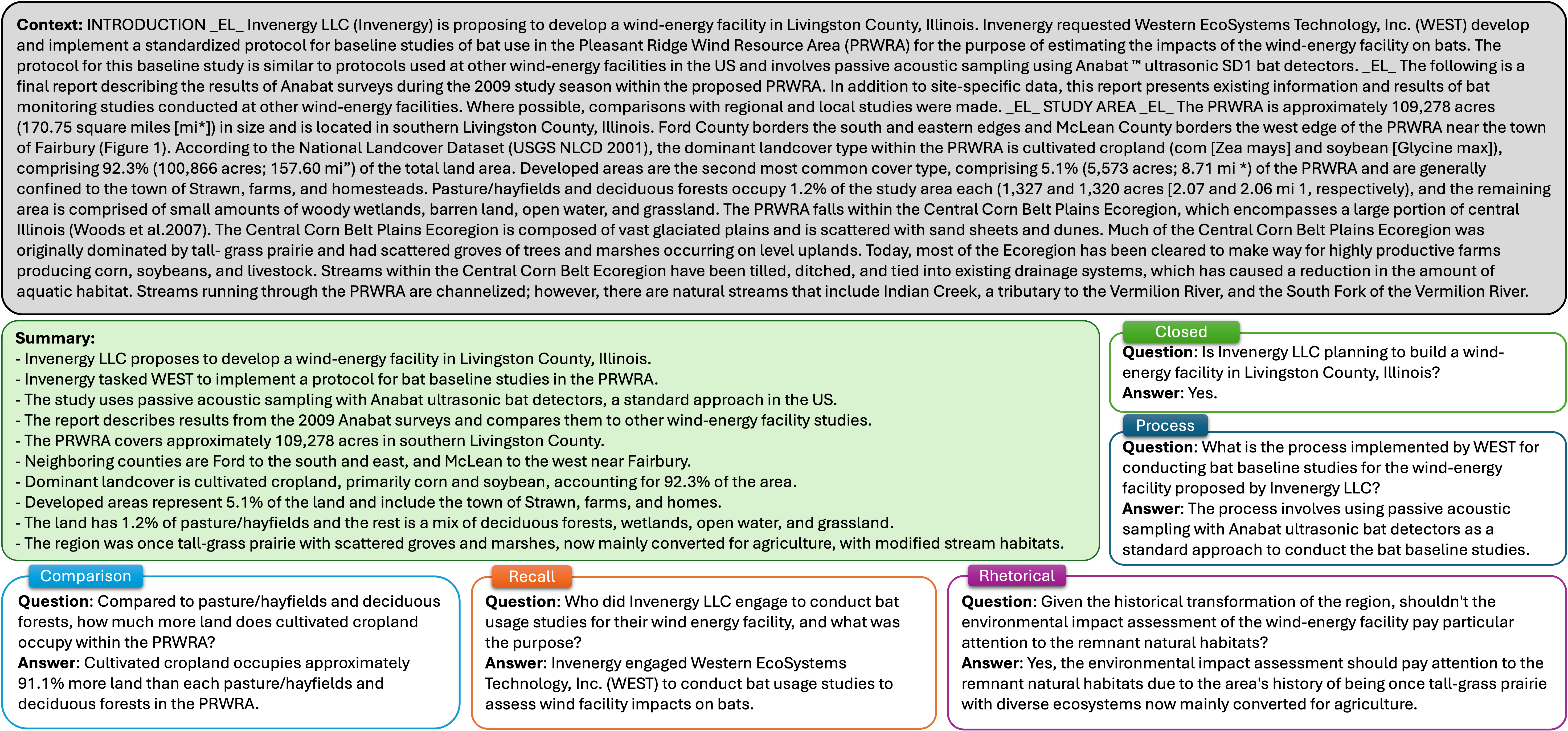}
    \caption{Different types of questions generated from the ``introduction'' section of a report~\cite{pleasant} generated by the \emph{hybrid context approach}. The section from the original document is first summarized and the question-answer pairs are generated from the summarized text chunk.}
    \label{fig:questions_all}
    \vspace{-1.5em}
\end{figure*}

We present two complementary approaches for hybrid question generation to support comprehensive \gls*{llm} benchmarking. 
The \emph{Hybrid Prompt Approach} employs engineered prompts to generate high-quality, curated questions, while the \emph{Hybrid Context Approach} leverages text summarization to create questions that require broader contextual understanding.
The detailed prompts used for question generation across both approaches are provided in the Appendix.

\noindent\textbf{Hybrid Prompt Approach.}~We utilize GPT-4 to automatically generate questions from given text chunks through carefully designed prompts tailored to each question type. 
To enhance question quality, we implement a manual curation process where domain experts identify exemplary questions that effectively assess LLM capabilities for benchmarking purposes. 
This curation is performed systematically across all question types, ensuring that each category incorporates appropriate grammatical structures and complexity levels. 
These curated questions subsequently serve as few-shot examples to guide the automatic question generation framework, improving the overall quality and consistency of generated questions.

\noindent\textbf{Hybrid Context Approach.}~The initial approach primarily generates questions at the sentence level by substituting subjects or objects with interrogative words, which proves adequate for `closed', `open', and `recall' type questions where answers can be directly extracted from the text.
However, `process', `evaluation', and `comparison' questions require deeper inferential reasoning across larger text segments. 
To address this limitation, we first employ GPT-4 to summarize extensive text chunks (typically exceeding 15 sentences) into concise summaries containing 5-8 sentences. 
We then generate questions from these summarized chunks using the hybrid prompt methodology combined with curated sample questions, ensuring that the resulting questions necessitate comprehensive understanding and synthesis of broader contextual information.





\noindent\textbf{Questions from tables.}~
An essential component of benchmarking \gls*{rag}-based \gls*{llm}s within research articles and reports involves evaluating their capability to retrieve and interpret tabular information. 
Tables represent critical content elements within research documents, frequently containing comprehensive summaries and key quantitative data that encapsulate the essence of entire sections or studies. 
To address this requirement, we extract tabular data in HTML format and systematically organize it within our JSON schema framework. 
This HTML-formatted tabular data is subsequently incorporated into our prompt engineering pipeline to generate targeted question-answer pairs that specifically assess the model's proficiency in understanding and reasoning over structured tabular information.

Figure~\ref{fig:questions_all} illustrates the diverse question-answer pairs generated from the introduction section of a document~\cite{pleasant} using our proposed methodology. 
We demonstrate the Hybrid Context approach where the section content is first summarized into a concise form, and subsequently, targeted QA pairs are generated from this summarized context to ensure comprehensive coverage of key concepts. 
Table~\ref{tbl:qt:distribution} presents the statistical distribution of different question types within the WeQA benchmark, providing insights into the composition and balance of our evaluation dataset.

\begin{table}
\centering
\caption{Question types in the WeQA benchmark}
\begin{tabular}{lrr}
\hline
\textbf{Type} & \textbf{\#Questions} &\textbf{\% Questions}   \\ \hline
Closed                                       & 382         & 18\% \\
Comparison                                   & 393            & 19\% \\ 
Evaluation                                   & 273          & 13\% \\
Rhetorical                                       & 324          & 16\% \\ 
Process                                      & 172         & 8\% \\
Recall                                       & 258         & 12\% \\ 
Open  &270 & 13\% \\
\hline
\end{tabular}
\label{tbl:qt:distribution}
\end{table}

\section{Results and Discussion}

\begin{figure*}
    \centering
    \includegraphics[width=0.48\textwidth]{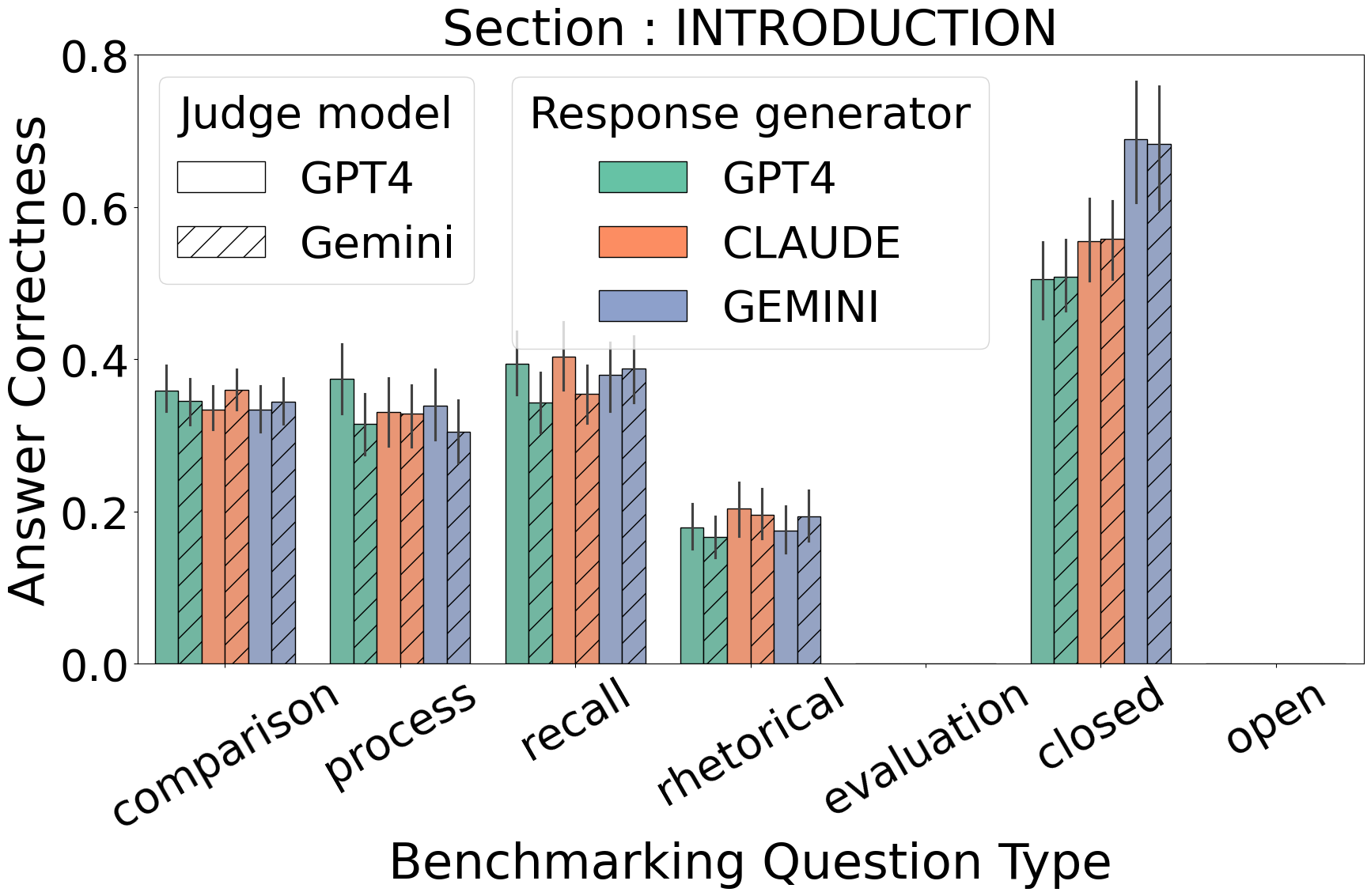}
    \includegraphics[width=0.48\textwidth]{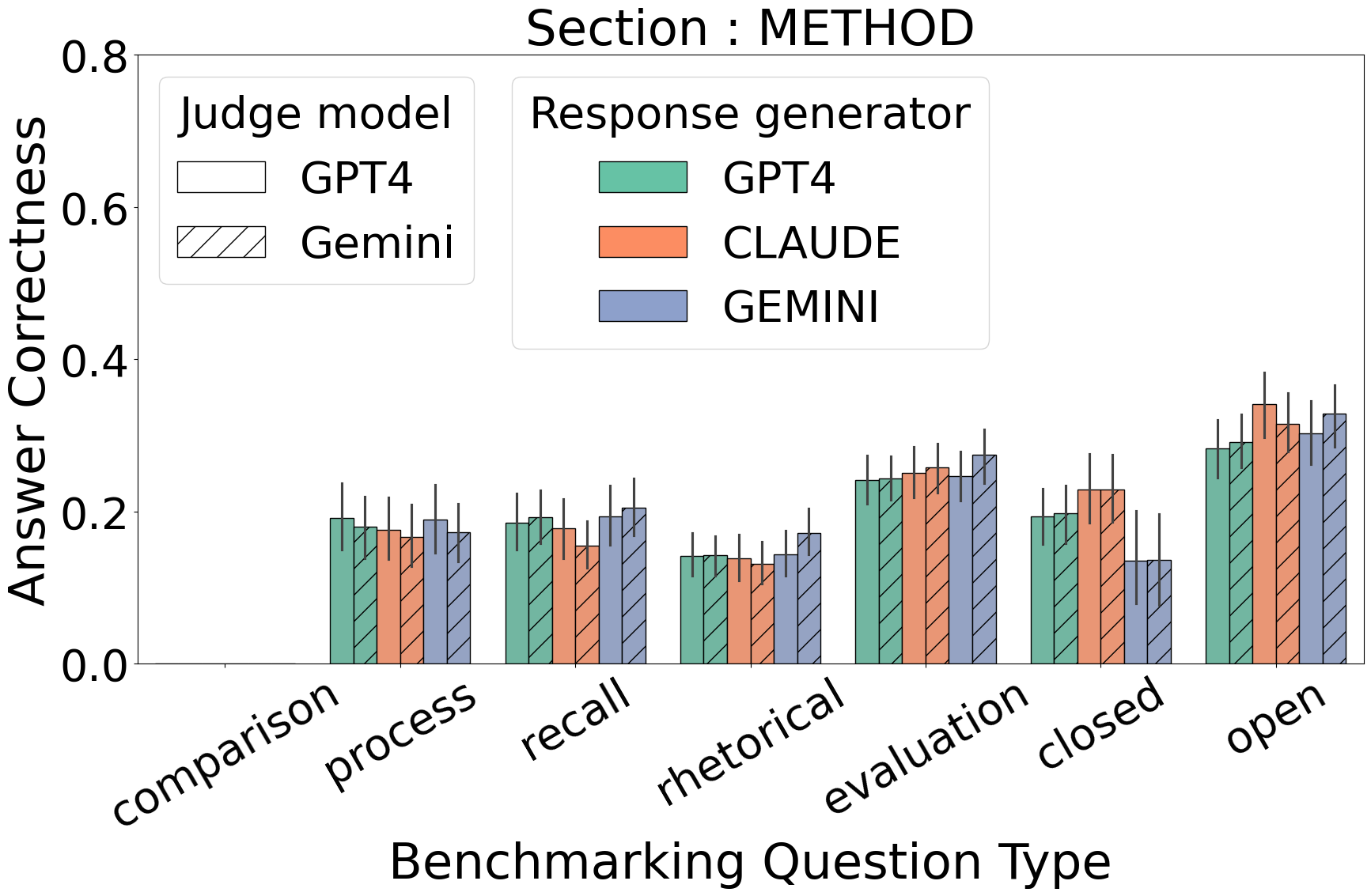}
    \includegraphics[width=0.48\textwidth]{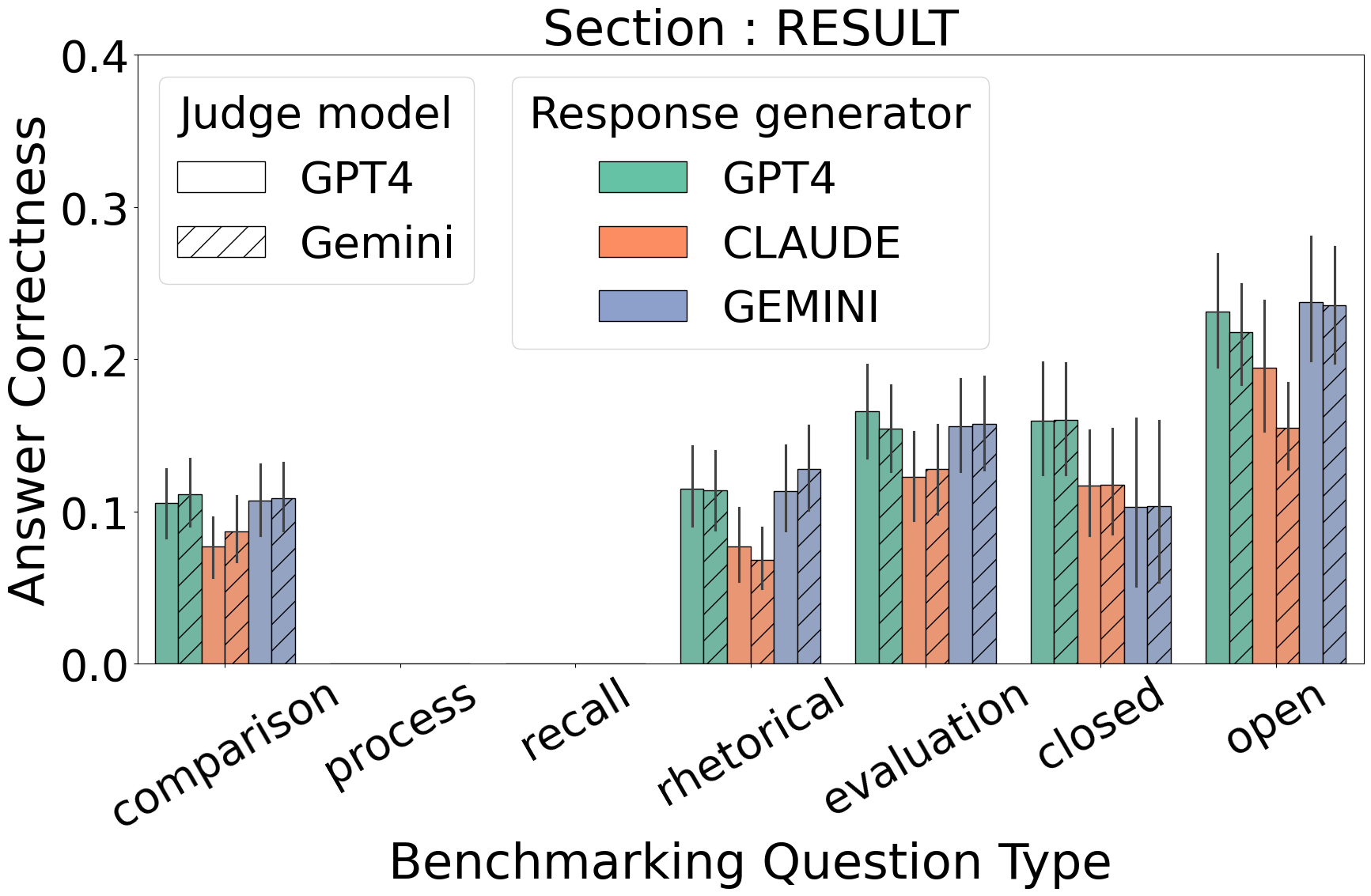}
    \includegraphics[width=0.48\textwidth]{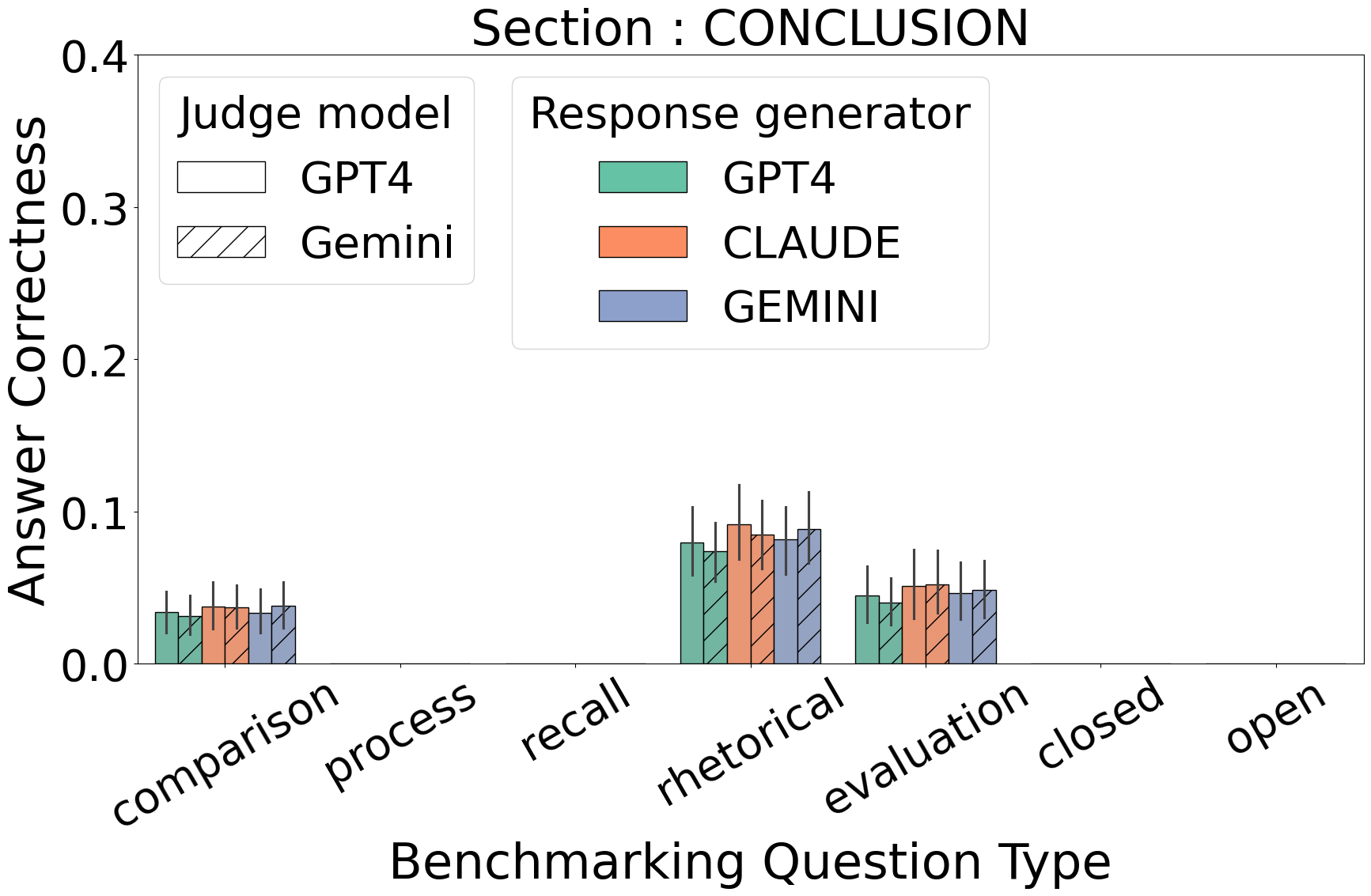}
    \caption{Answer correctness scores computed using the RAGAS scoring framework with GPT-4 and Gemini-1.5Pro as judge models for response generated by all three models used.}
    \label{fig:answer-score}
\end{figure*}

\textbf{Experimental setup.}~We conduct a comprehensive evaluation of three state-of-the-art LLMs—GPT-4, Gemini, and Claude—on our WeQA benchmark within a RAG framework
Knowledge extraction is performed from wind energy documents to create vector embeddings as shown in the data-curation pipeline in Figure~\ref{fig:framework}, which are subsequently stored in a vector database to enable retrieval-augmented generation capabilities.
We employ the \ragas evaluation framework, leveraging judge LLMs to provide systematic assessment of model performance across multiple dimensions. The evaluation encompasses key metrics including answer correctness, context precision, and context recall, offering comprehensive insights into each model's proficiency in both retrieving relevant information and generating accurate responses from the provided context. 
For the judge LLM component, we utilize both GPT-4 and Gemini-1.5Pro to ensure robust and unbiased evaluation of the assessed models' performance.
Figure~\ref{fig:answer-score} presents the answer correctness score, while the context precision and context recall depicted in Table~\ref{tab:appendix-results} (added in Appendix) show the ability of the models to retrieve the context accurately.

\begin{observation}
The observed answer correctness scores are notably low, indicating a robust and challenging benchmark.
\end{observation}
Specifically, "evaluation" and "comparison" type questions yield nearly zero answer correctness scores for all models, highlighting their difficulty in responding. Recall that, these challenging questions were crafted from summaries of text chunks rather than the text chunks themselves, further complicating the models' ability to generate correct answers. This underscores the complexity and rigor of the benchmarking process, emphasizing the need for models to improve their understanding and contextual extraction capabilities.

\begin{observation}
There is an alignment in evaluations made by the two judge \gls*{llm}s used within the \ragas framework, particularly visible for `closed' type questions. 
\end{observation}
This similarity arises because the answers to these questions are objective (`yes' or `no'), leading to equivalent correctness evaluations by both models.
Although there are some mismatches in the evaluations made by the two judge \gls*{llm}s, the number of these discrepancies is insignificant compared to the number of matching evaluations.

\begin{figure}[ht]
    \centering
    \includegraphics[width=0.48\textwidth]{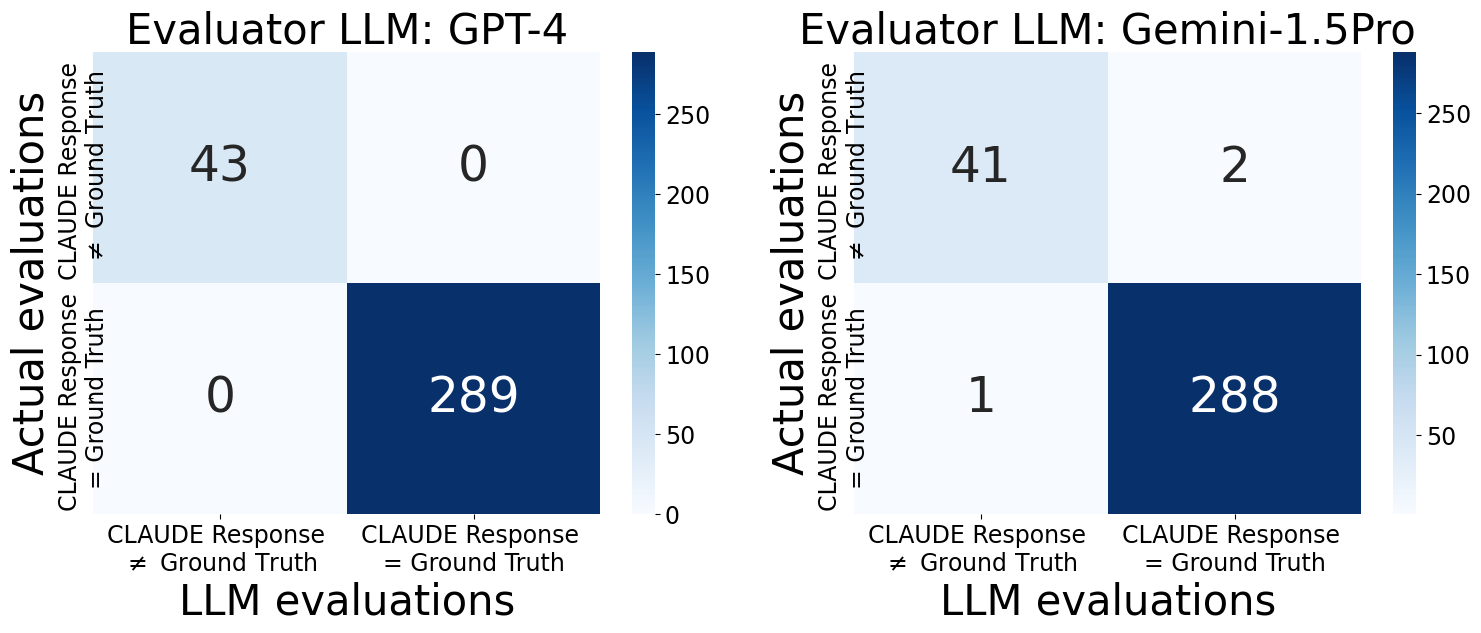}
    \includegraphics[width=0.48\textwidth]{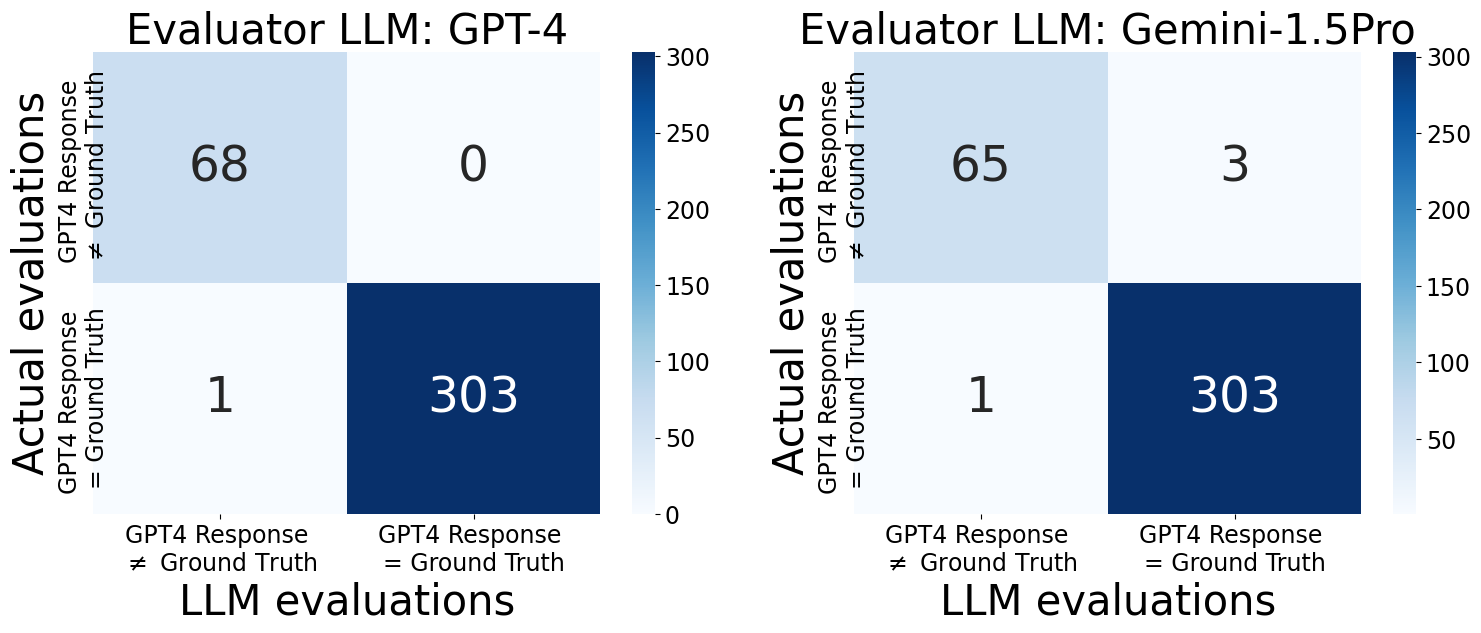}
    \caption{Confusion Matrix for evaluations by judge \gls*{llm}s on responses from Claude (top) and GPT-4 (bottom) models}
    \label{fig:confusion-mat}
\end{figure}

Figure~\ref{fig:confusion-mat} displays the confusion matrix illustrating the evaluations made by the two judge \gls*{llm}s (GPT-4 and Gemini-1.5Pro) on the responses provided by the \gls*{rag}-based Claude and GPT-4 models to the benchmarking questions. 
In this context, a true positive occurs when the judge \gls*{llm} correctly identifies the model response as matching the ground truth. Conversely, a false positive arises when the judge \gls*{llm} incorrectly states that the model response matches the ground truth, while it does not. This matrix helps visualize the accuracy and reliability of the evaluations conducted by the \gls*{llm}s, when used within the \ragas framework.
We note that majority of evaluations made by either judge \gls*{llm} matches the actual evaluation which indicates that both of them are reliable.

\begin{observation}
Comparison between `closed' and `open' type questions within the same section reveals a higher answer correctness for responses to `open' type questions than `closed' type questions.
\end{observation}
From this observation, we conclude that \gls*{rag}-based models generate more accurate subjective responses to `open' questions than objective (`yes' or `no') responses for `closed' questions. 
This phenomenon may stem from the inherent design of \gls*{llm}s, which are optimized for generating extensive text sequences and may struggle with the precision required for definitive binary responses. 
This suggests that these models perform better when tasked with generating detailed, context-rich answers rather than simple, binary ones, highlighting their strength in handling nuanced and complex queries.

\begin{observation}
The answer correctness scores for questions derived from the ``Introduction'' section are higher compared to those from other sections.
\end{observation}
This is because the ``introduction'' section is typically longer, more similar to other documents, and often includes a related works section, which aligns closely with content found in many other documents. 
As a result, the \gls*{rag}-based \gls*{llm}s can more easily extract relevant information to answer questions accurately, leading to higher correctness scores. 
Additionally, the content in the ``introduction'' section is primarily text-based, unlike other sections which contain equations, tables, and figures. Therefore, the models provide more accurate responses to questions from the ``introduction'' section compared to those from other sections.

\begin{observation}
The answer correctness scores for `rhetorical' questions are lower than those for other question types.
\end{observation}
This is because `rhetorical' questions lack definite answers in the documents, making it challenging for the models to retrieve the appropriate context and provide correct responses. 
The absence of clear, concrete answers in the source material complicates the models' ability to generate accurate and relevant responses, leading to lower correctness scores for this question type.

\begin{figure}[t]
    \centering
    \includegraphics[width=0.235\textwidth]{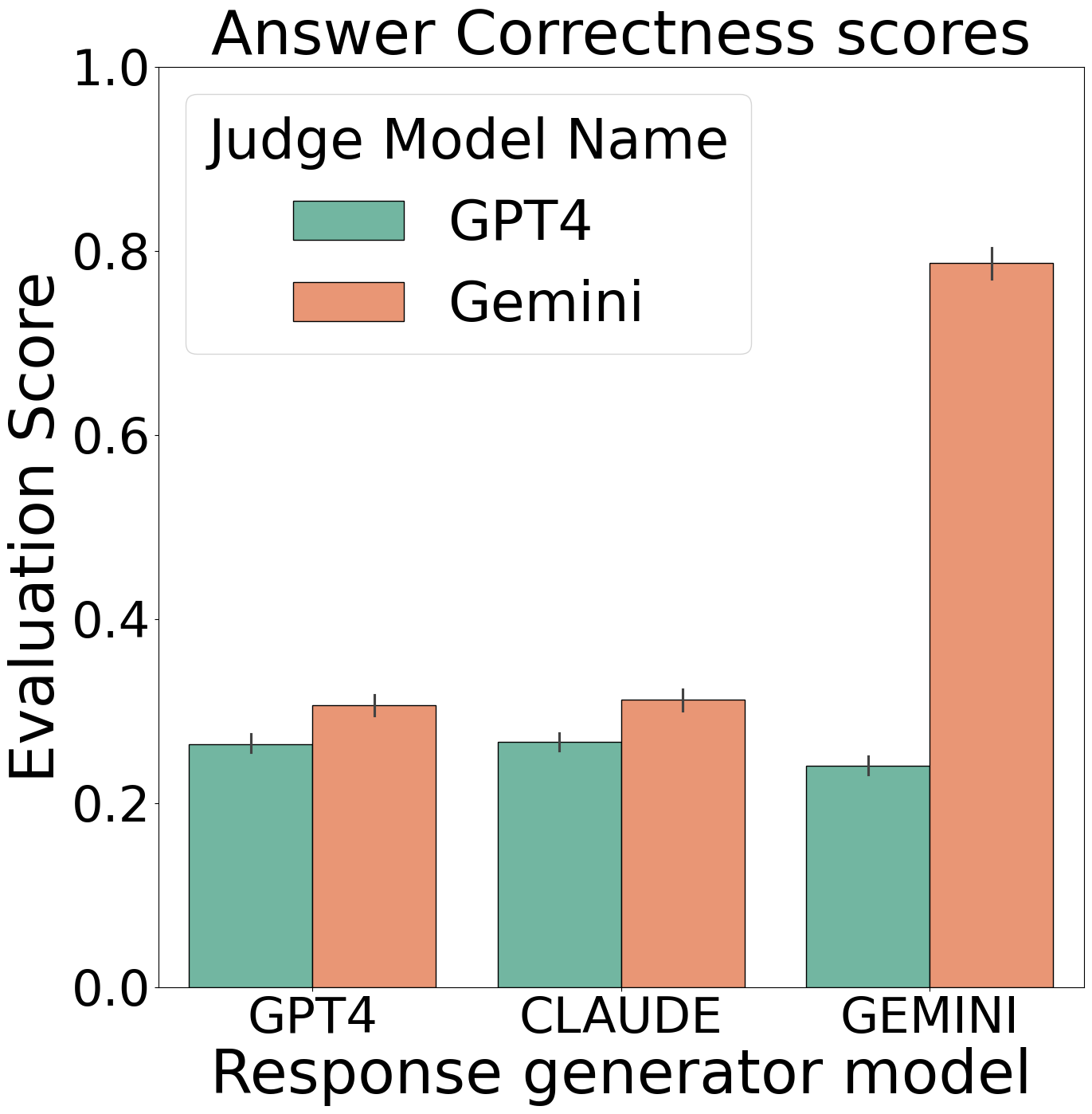}
    \includegraphics[width=0.235\textwidth]{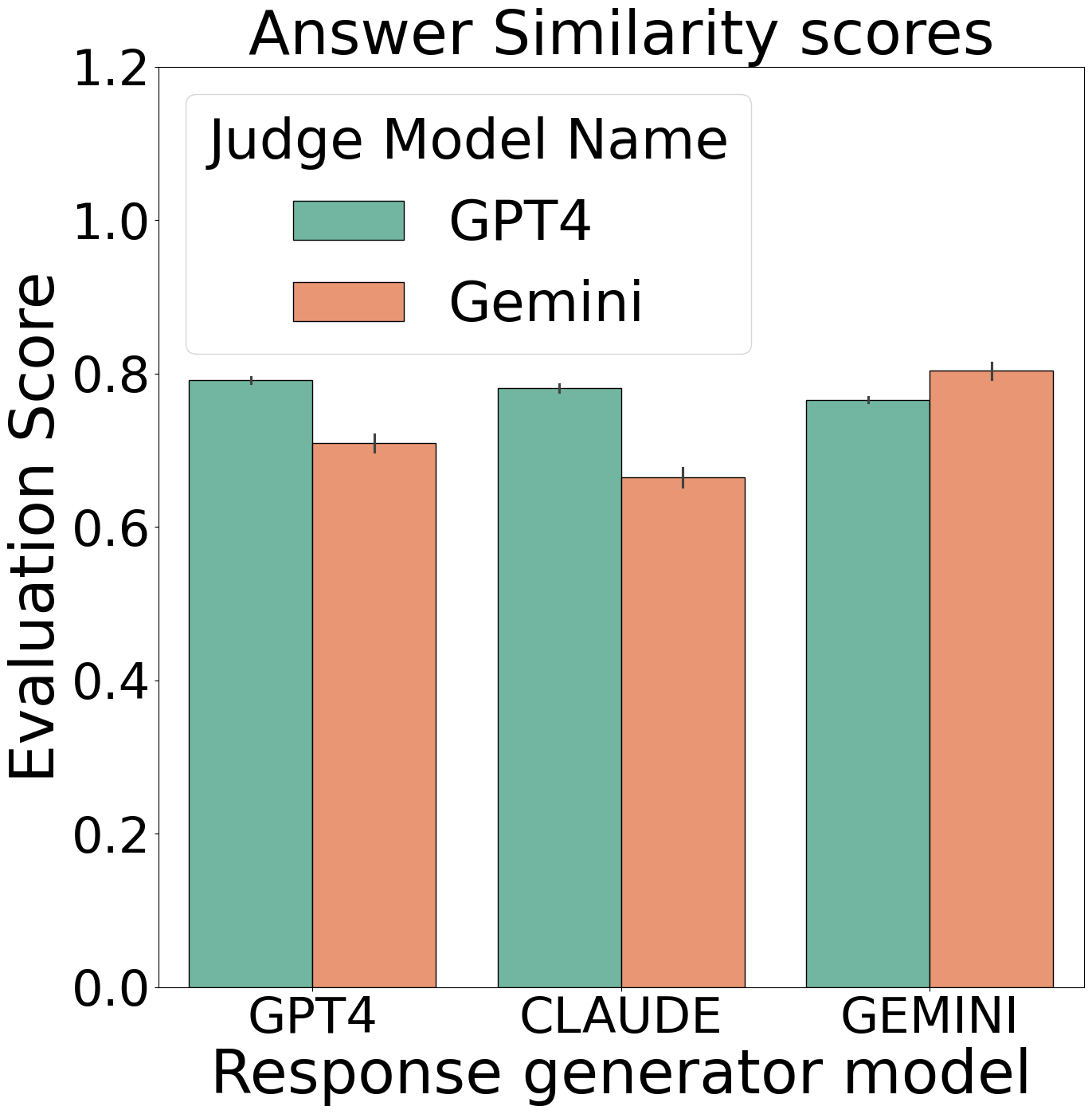}
    \includegraphics[width=0.235\textwidth]{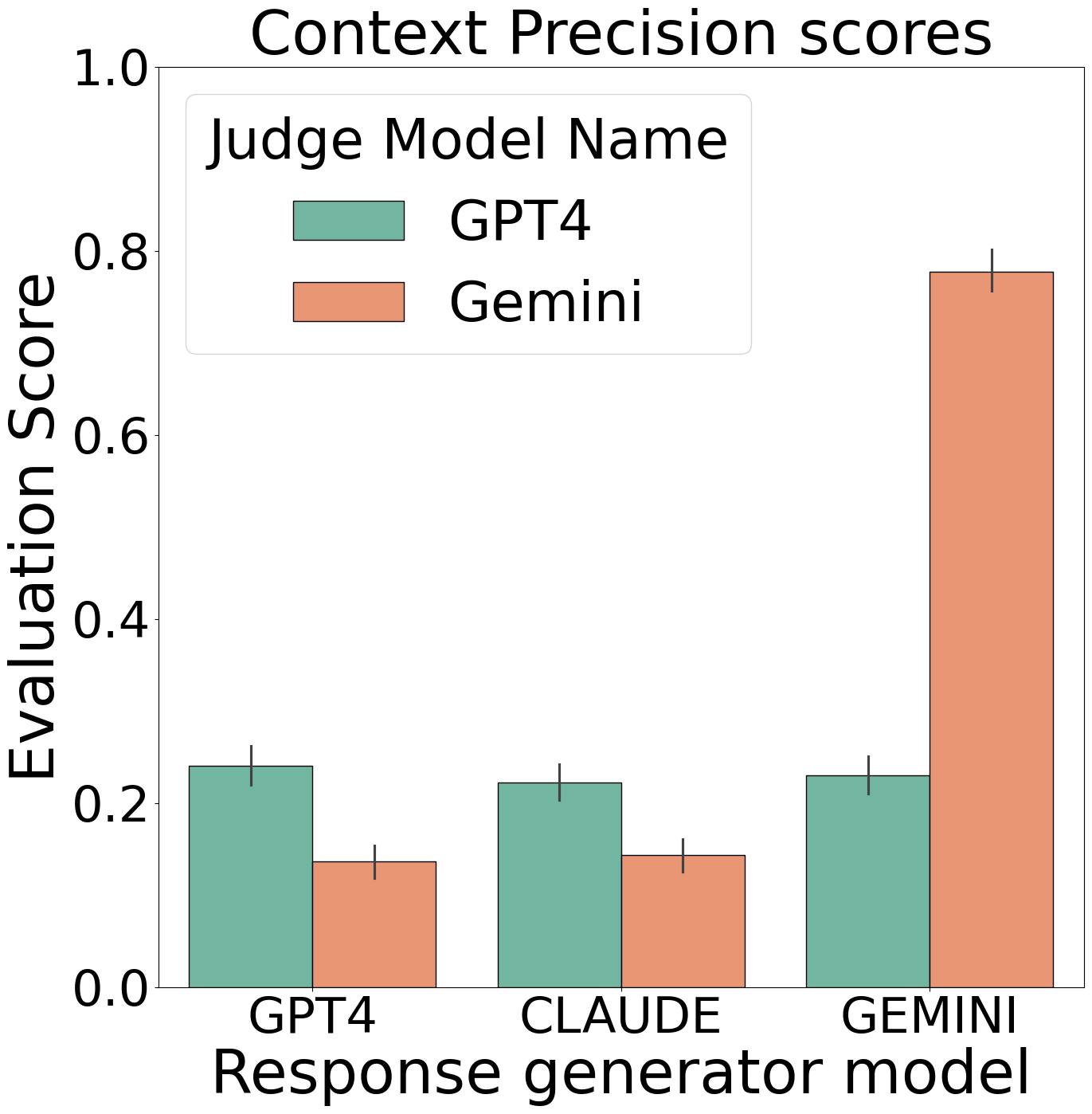}
    \includegraphics[width=0.235\textwidth]{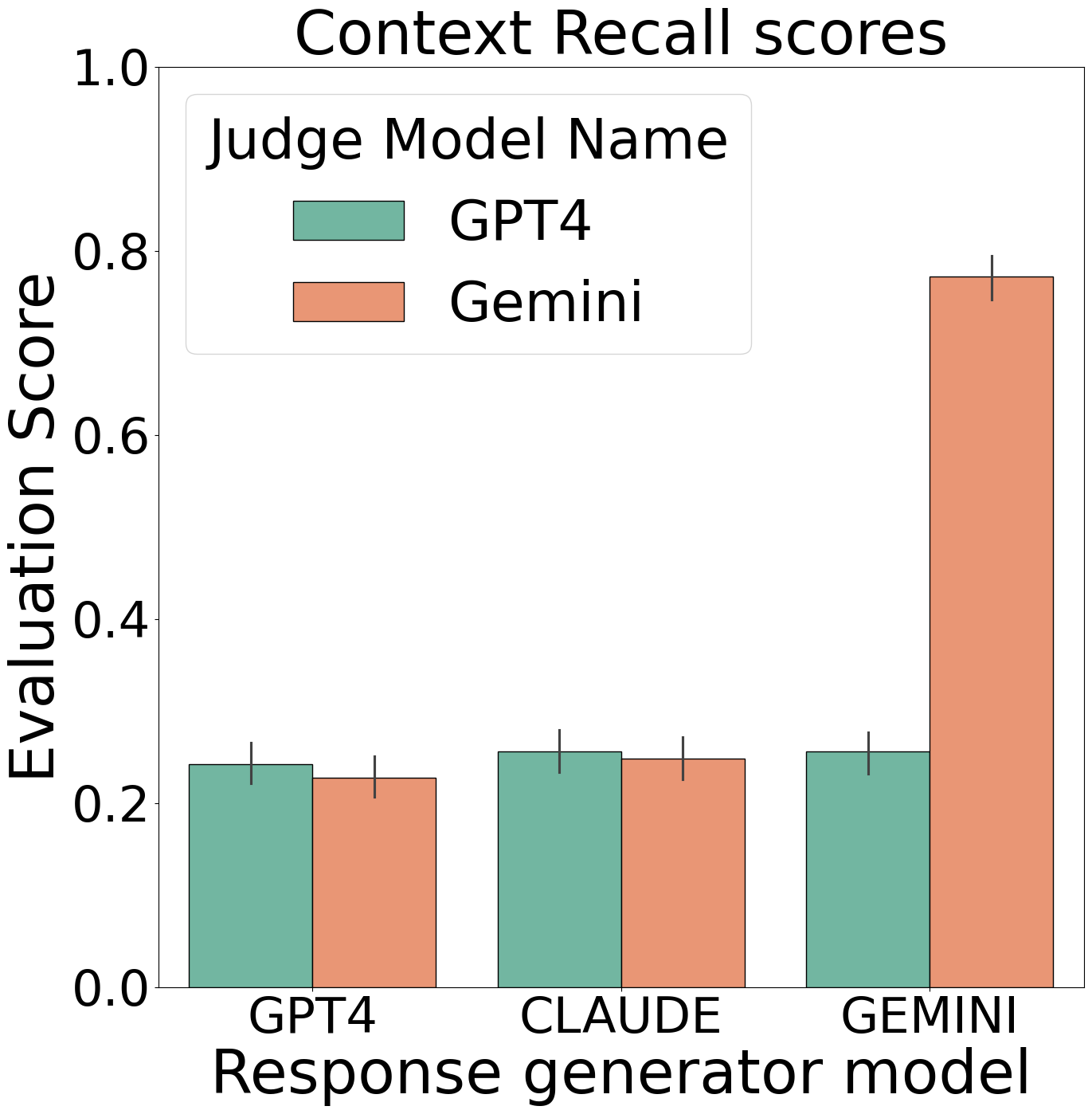}
    \caption{Answer correctness (top left), answer similarity (top right), context precision (bottom left) and recall (bottom right) scores across different judge and generator models.}
    \label{fig:table-answer-score}
\end{figure}

\begin{observation}
Evaluations made by Gemini-1.5Pro on the responses generated by all three \gls*{llm}s are higher than the evaluations made by GPT-4, with the responses from Gemini \gls*{llm} receiving significantly higher scores.
\end{observation}
Figure~\ref{fig:table-answer-score} shows the scores computed by the judge \gls*{llm}s for the responses generated by the three \gls*{rag}-based \gls*{llm}s. The Gemini-1.5Pro judge tends to rate high scores even when the \gls*{llm}s refuse to answer. An example is listed below:
\eval{\textbf{Question:}~In the HTML table that estimates the annual number of bird collisions at different percentages of avoidance, what is the estimated number of collisions per year at $98.0\%$ avoidance?\\
\textbf{Expected answer:}~The estimated number of collisions per year at $98.0\%$ avoidance is $152$.\\
\textbf{Gemini generated answer:}~This question cannot be answered from the given source. While the text discusses bird collisions with wind turbines, it does not provide any data on estimated collision numbers at different avoidance rates.\\
\textbf{GPT-4 score:}~0.194\\
\textbf{Gemini-1.5Pro score}~0.813}

Conversely, in instances where the LLMs generate correct answers, Gemini-1.5Pro has been observed to evaluate them as incorrect; such as:
\eval{\textbf{Question:}~Who is the GIS Technician in the `STUDY PARTICIPANTS' table?\\
\textbf{Expected answer:}~JR Boehrs\\
\textbf{Gemini generated answer:}~Saif Nomani JR Boehrs was the GIS Technician.\\
\textbf{GPT-4 score:}~0.703\\
\textbf{Gemini-1.5Pro score:}~0.200}

\section{Conclusion}
\label{sec:conclusion}
In conclusion, this paper presents a versatile framework for evaluating the performance of \gls*{rag}-based \gls*{llm}s across various question types and document sections. We showcase this by introducing a hybrid, automated question-generation method that ensures comprehensive coverage of both objective and subjective queries, and implement this for the use case of wind energy related document and present the WeQA benchmark, which is a first of its kind benchmark in wind energy domain. However, the usefulness of our work goes beyond this niche domain as our approach is domain-agnostic, meaning it can be used for creating benchmark for any domain. Additionally, our use of the RAGAS scoring framework allows for a thorough evaluation of model performance, offering a holistic assessment of LLM capabilities, while also having the advantage of being easy for other researchers to adapt this approach for their own work.

\section{Limitations}
\label{sec:limitations}
A limitation of the proposed framework is that the automatic method of generating questions often produces queries that are too specific to the document from which they were derived. When these questions are posed to an \gls*{llm} with a large document corpus, the model may struggle to respond accurately, necessitating the filtering of ambiguous questions to ensure relevance and clarity. Additionally, the \ragas scoring framework, which relies on \gls*{llm}s as judges, introduces uncertainty in performance metrics, as different judge \gls*{llm}s may score responses differently. While comparisons can be made for questions with objective responses, evaluating and comparing subjective responses across different \gls*{llm}s remains challenging and less consistent.
Another limitation of this study is the absence of comprehensive ablation studies, including comparisons between \gls*{rag}-enabled and non-\gls*{rag} configurations, which would provide deeper insights into the specific contributions of retrieval mechanisms to model performance.

\section{Ethical Considerations}
\label{sec:ethics}
While we do not anticipate the novel work presented here to introduce new ethical concerns in and by themselves, we do recognize that there may also be pre-existing concerns and issues of the data, models, and methodologies we have used for this paper. We acknowledge that researchers should not “simply assume that [...] research will have a net positive impact on the world” \cite{hecht2021s}. In particular, it has been seen that Large Language Models (LLMs), like the ones used in this work, exhibit a wide variety of bias -- \textit{e.g.,~}religious, gender, race, profession, and cultural -- and frequently generate answers that are incorrect, misogynistic, antisemitic, and generally toxic \cite{abid2021persistent,buolamwini2018gender,liang2021towards,nadeem-etal-2021-stereoset,welbl2021challenges}.
However, when used within the parameters of our experiments detailed in this paper, we did not see such behaviour from any of the models. To our knowledge, when used as intended, our models do not pose additional ethical concerns than any other LLM.


\bibliography{main}

\appendix
\section{Prompts used to generate QA pairs using Hybrid Prompt Approach}
In this section, we detail the various prompts used to create the different types of questions in the WeQA benchmark dataset.
First, we show the prompt to generate questions from a given text chunk. We use curly braces to denote placeholders for the different inputs to the prompt.
\prompt{Generate \{\textbf{number of questions}\} questions given the content provided in the following paragraph. Restrict the type of questions to \{\textbf{question type}\} questions.\\
\{\textbf{Text chunk from document section}\}}

We curate the generated questions, where domain experts manually identify the questions which are best suited for the purpose of benchmarking \gls*{llm}s.
We perform this process for each type of question, so that we include particular grammatical structures for each question type.
Thereafter, we use these curated high-quality questions as \emph{few-shot examples} to regenerate questions using the automatic question generation framework.
The updated prompt along with the placeholders looks as follows:
\prompt{Generate \{\textbf{number of questions}\} questions given the content provided in the following paragraph. Restrict the type of questions to \{\textbf{question type}\} questions.\\
\{\textbf{Text chunk from document section}\}\\
You can generate similar questions (but not limited) to sample questions provided below.\\
\{\textbf{Sample question 1}\}\\\{\textbf{Sample question 2}\}\\\{\textbf{Sample question 3}\}}

\section{Prompts used to generate QA pairs using Hybrid Context Approach}
We use the following prompt to summarize a document section from which the questions are to be generated.
\prompt{You are a smart assistant. Can you summarize this input paragraph within \{\textbf{number of points}\} bullet points. Return the summarized text.\\
Input paragraph:
\{\textbf{Text chunk from document to summarize}\}
}
Thereafter, we use the earlier prompt to generate questions from this summarized text chunk. We add the few-shot example questions which are identified by the domain experts for each question type.
\prompt{Generate \{\textbf{number of questions}\} questions given the content provided in the following paragraph. Restrict the type of questions to \{\textbf{question type}\} questions.\\
\{\textbf{Summarized text chunk from document section}\}\\
You can generate similar questions (but not limited) to sample questions provided below.\\
\{\textbf{Sample question 1}\}\\\{\textbf{Sample question 2}\}\\\{\textbf{Sample question 3}\}}

\section{Prompts used to generate QA pairs from tables}
We extract the tabular data from documents as HTML objects in the filtered JSON schema.
We use the following prompt to generate question-answer pairs from the tabular data.
\prompt{Generate \{\textbf{number of questions}\} questions given the table provided in HTML format in the following paragraph?
Generate the questions keeping in mind that the caption of the table is \{\textbf{Table caption obtained from document.}\}\\
Restrict the questions such that the answers can be retrieved from the provided table in the HTML format.
For each question, return 3 lines: question/ answer/ proof.
Make sure there are no newline characters in the proof.\\
Input table:\{\textbf{Table in HTML format extracted from document}\}
}

We show an example QA pair generated from a table obtained from a document~\cite{pleasant}.
Table~\ref{tab:tethys-table} shows the table from the document for reference.
An example QA-pair generated from this table is provided here.
\qa{\textbf{Question}: What is the acreage of Cultivated Crops within the Pleasant Ridge Project Area based on the National Land Cover Database in May of 2014? \\\textbf{Answer}: The acreage of Cultivated Crops within the Pleasant Ridge Project Area is 55,946 acres.\\\textbf{Proof:}~The table entry under the “Habitat” column for ``Cultivated Crops'' corresponds with the entry under the ``Acres [Hectares]'' column that reads ``55,946[22,641]}
\begin{table}
\centering
\caption{\small{Land Cover Types, Coverage, and Composition within the Pleasant Ridge Project Area, Based on National Land Cover Database in May of 2014~\cite{pleasant}}}
\label{tab:tethys-table}
\begin{tabular}{lcc}
\hline
\small{\textbf{Habitat}}                        & \small{\textbf{Acres {[}Hectares{]}}} & \small{\textbf{\% Composition}} \\ \hline
\small{Cultivated Crops}                        & \small{55,946{[}22,641{]}}            & \small{92.6}                    \\
\small{Developed}                               & \small{3,432{[}1,389{]}}              & \small{5.7}                     \\
\small{Deciduous Forest}                        & \small{451{[}183{]}}                  & \small{0.7}                     \\
\small{Hay/Pasture}                             & \small{347{[}140{]}}                  & \small{0.6}                     \\
\small{Open Water}                              & \small{122{[}49{]}}                   & \small{0.2}                     \\
\small{Woody Wetlands}                          & \small{111{[}45{]}}                   & \small{0.2}                     \\
\small{Barren Land}                             & \small{19{[}8{]}}                     & \small{0.0}                     \\
\small{Herbaceous}                              & \small{3{[}1{]}}                      & \small{0.0}                     \\ \hline
\small{\textbf{Total}}                          & \small{\textbf{60,431{[}24,456{]}}}   & \small{\textbf{100}}            \\ \hline
\end{tabular}
\end{table}

\section{Context Recall and Context Precision}
We utilize \ragas context recall and precision metrics to evaluate the retrieval performance of our \gls*{rag}-based systems, where context recall measures the proportion of relevant information successfully retrieved from the knowledge base, and context precision assesses the relevance of the retrieved context to the given query.
In our setup, we employ semantic similarity-based retrieval using vector embeddings, where `relevant context' is defined as text chunks or the document sections that contain information necessary to answer the posed questions.
\begin{table*}
    \centering
    \small
    \begin{tabular}{ll|rrrrrr|rrrrrr}
    \toprule
     &  & \multicolumn{6}{c}{GPT-4 as Judge} & \multicolumn{6}{c}{Gemini 1.5 Pro as Judge} \\
     & \underline{Model $\rightarrow$ } & \multicolumn{2}{c}{\underline{\hspace{5pt}GPT\hspace{5pt}}} & \multicolumn{2}{c}{\underline{\hspace{5pt}Claude\hspace{5pt}}} & \multicolumn{2}{c|}{\underline{\hspace{5pt}Gemini\hspace{10pt}}} & \multicolumn{2}{c}{\underline{\hspace{5pt}GPT\hspace{5pt}}} & \multicolumn{2}{c}{\underline{\hspace{5pt}Claude\hspace{5pt}}} & \multicolumn{2}{c}{\underline{\hspace{5pt}Gemini\hspace{10pt}}}\\
    Section $\downarrow$ & Type $\downarrow$ & Prec. & Rec. & Prec. & Rec. & Prec. & Rec. & Prec. & Rec. & Prec. & Rec. & Prec. & Rec. \\
    \midrule
    \multirow{5}{*}{Introduction} & closed & 0.467 & 0.314 & 0.500 & 0.330 & \textbf{0.570} & \textbf{0.385} & 0.392 & 0.435 & 0.424 & 0.448 & \textbf{0.467} & \textbf{0.563}\\
    & comparison & 0.556 & 0.596 & \textbf{0.607} & \textbf{0.672} & 0.587 & 0.628 & 0.429 & 0.597 & \textbf{0.480} & \textbf{0.637} & 0.454 & 0.632 \\
    & process & 0.565 & 0.608 & \textbf{0.598} & \textbf{0.625} & 0.586 & 0.602 & 0.457 & 0.568 & 0.467 & 0.603 & \textbf{0.483} & \textbf{0.591}\\
    & recall & 0.529 & 0.597 & \textbf{0.560} & \textbf{0.617} & 0.540 & 0.586 & 0.491 & 0.611 & \textbf{0.487} & \textbf{0.624} & 0.483 & 0.601 \\
    & rhetorical & 0.305 & 0.296 & \textbf{0.365} & \textbf{0.353} & 0.319 & 0.306 & 0.272 & 0.299 & \textbf{0.323} & \textbf{0.339} & 0.283 & 0.299\\
    \midrule
    \multirow{6}{*}{Method} & closed & 0.162 & 0.119 & \textbf{0.168} & \textbf{0.139} & 0.094 & 0.082 & 0.128 & 0.176 & \textbf{0.144} & \textbf{0.174} & 0.084 & 0.093\\
    & open &0.364 & 0.431 & \textbf{0.431} & \textbf{0.540} & 0.378 & 0.471 & 0.333 & 0.455 & \textbf{0.383} & \textbf{0.511} & 0.367 & 0.446\\
    & evaluation &0.400 & 0.387 & \textbf{0.442} & \textbf{0.453} & 0.416 & 0.422 & 0.311 & 0.406 & \textbf{0.352} & \textbf{0.474} & 0.316 & 0.430\\
    & process &0.270 & 0.275 & 0.270 & 0.293 & \textbf{0.282} & \textbf{0.302} & 0.209 & 0.282 & 0.162 & 0.268 & \textbf{0.210} & \textbf{0.306} \\
    & recall &0.234 & 0.277 & 0.223 & 0.268 & \textbf{0.250} & \textbf{0.285} & \textbf{0.223} & \textbf{0.270} & 0.188 & 0.251 & 0.212 & 0.278\\
    & rhetorical &0.229 & 0.223 & 0.241 & 0.232 & \textbf{0.250} & \textbf{0.238} & 0.208 & 0.238 & 0.193 & 0.230 & \textbf{0.224} & \textbf{0.248}\\
    \midrule
    \multirow{5}{*}{Results} & closed & \textbf{0.143} & \textbf{0.077} & 0.102 & 0.072 & 0.076 & 0.059 & \textbf{0.120} & \textbf{0.101} & 0.093 & 0.099 & 0.070 & 0.086 \\
     & open &0.284 & 0.328 & 0.263 & 0.280 & \textbf{0.325} & \textbf{0.320} & 0.230 & 0.306 & 0.192 & 0.265 & \textbf{0.253} & \textbf{0.320} \\
     & comparison &0.167 & 0.174 & 0.139 & 0.141 & \textbf{0.172} & \textbf{0.173} & 0.128 & 0.157 & 0.098 & 0.119 & \textbf{0.134} & \textbf{0.156} \\
     & evaluation &\textbf{0.272} & \textbf{0.254} & 0.217 & 0.218 & 0.257 & 0.263 & \textbf{0.226} & \textbf{0.252} & 0.171 & 0.229 & 0.209 & 0.266 \\
     & rhetorical &\textbf{0.192} & \textbf{0.182} & 0.133 & 0.126 & 0.183 & 0.175 & 0.156 & 0.180 & 0.100 & 0.136 & \textbf{0.160} & \textbf{0.176} \\
    \midrule
     \multirow{3}{*}{Conclusion} & comparison & 0.048 & 0.051 & \textbf{0.059} & \textbf{0.065} & 0.055 & 0.058 & 0.045 & 0.050 & \textbf{0.053} & \textbf{0.059} & 0.050 & 0.058 \\
     & evaluation &0.082 & 0.079 & \textbf{0.100} & \textbf{0.103} & 0.086 & 0.089 & 0.073 & 0.081 & 0.072 & 0.084 & \textbf{0.078} & \textbf{0.081} \\
     & rhetorical &0.138 & 0.141 & \textbf{0.178} & \textbf{0.171} & 0.148 & 0.147 & 0.126 & 0.148 & \textbf{0.149} & \textbf{0.165} & 0.133 & 0.144 \\
    \midrule
    \bottomrule
    \end{tabular}
    \caption{Performance of the models on the WeQA benchmark scored using the RAGAS framework across judge \gls*{llm}s. The "Prec." and "Rec." mean Context Precision and Context Recall respectively, while "Type" refers to the Question Type. The best performance for each question type per judge \gls*{llm} is highlighted in bold.}
    \label{tab:appendix-results}
\end{table*}

\section{Judge LLM Evaluation Analysis Through Confusion Matrices}
To assess the reliability and accuracy of \gls*{llm}s as judges within the \ragas evaluation framework, we conduct a detailed analysis using confusion matrices for closed-type questions where binary (`yes'/`no') responses can be objectively compared against ground truth answers.
This analysis is particularly crucial for validating the trustworthiness of automated evaluation systems in benchmarking scenarios.

\noindent\textbf{Methodology for evaluation.}~We evaluate two judge \gls*{llm}s—GPT-4 and Gemini-1.5Pro—by comparing their assessments of \gls*{rag}-based model responses (Claude and GPT-4) against manually verified ground truth labels for closed-type questions. The confusion matrix framework allows us to quantify four key evaluation scenarios:
\begin{itemize}
    \item \textbf{True Positive (TP)}: The judge LLM correctly identifies that the model response matches the ground truth answer.
    \item \textbf{False Positive (FP)}: The judge LLM incorrectly states that the model response matches the ground truth when it does not
    \item \textbf{True Negative (TN)}: The judge LLM correctly identifies that the model response does not match the ground truth answer
    \item \textbf{False Negative (FN)}: The judge LLM incorrectly states that the model response does not match the ground truth when it actually does
\end{itemize}

\noindent\textbf{Analysis of judge LLM performance.}~The confusion matrices reveal that the majority of evaluations made by both judge LLMs align with the actual ground truth evaluations, demonstrating their reliability as automated evaluators. 
Specifically, both GPT-4 and Gemini-1.5Pro exhibit high accuracy rates in distinguishing correct from incorrect responses, with minimal discrepancies in their assessment capabilities.

\noindent\textbf{Cross-judge agreement.}~Additionally, we observe substantial agreement between the two judge \gls*{llm}s, suggesting consistency in evaluation standards across different model architectures. 
This cross-validation approach enhances the robustness of our evaluation methodology and provides confidence in the reliability of automated assessment within specialized domain benchmarks like WeQA.
\end{document}